\newcommand{\etal}{et al.}

\documentclass[conference]{IEEEtran}
\IEEEoverridecommandlockouts

\usepackage{amsmath,amssymb,amsfonts}
\usepackage{algorithmic}
\usepackage{graphicx}
\usepackage{textcomp}
\usepackage{xcolor}

\usepackage{bbding}
\usepackage{latexsym}
\usepackage{booktabs}
\usepackage{enumitem}
\usepackage{graphicx}
\usepackage{wrapfig}
\usepackage{color}
\usepackage{subfigure}
\usepackage{framed}
\usepackage{caption}
\usepackage{bigstrut}
\usepackage{multirow}
\usepackage{hyperref} 
\usepackage[utf8]{inputenc}
\usepackage{newunicodechar}
\definecolor{shadecolor}{rgb}{0.92,0.92,0.92}
\hypersetup{
    colorlinks=true,
    linkcolor=red,
    citecolor=blue,
}
\newunicodechar{Ⅰ}{I}
\newunicodechar{Ⅱ}{II}

\def\BibTeX{{\rm B\kern-.05em{\sc i\kern-.025em b}\kern-.08em
    T\kern-.1667em\lower.7ex\hbox{E}\kern-.125emX}}
\begin{document}

\title{Who is Undercover? Guiding LLMs to Explore Multi-Perspective Team Tactic in the Game}

\author{
Ruiqi Dong\textsuperscript{1},
Zhixuan Liao\textsuperscript{1},
Guangwei Lai\textsuperscript{1},
Yuhan Ma\textsuperscript{1},
Danni Ma\textsuperscript{1},
Chenyou Fan\textsuperscript{1,*}\\
\textsuperscript{1}South China Normal University, Guangzhou, China\\
2023025193@m.scnu.edu.cn, fanchenyou@scnu.edu.cn
}

\maketitle

\begin{abstract}
Large Language Models (LLMs) are pivotal AI agents in complex tasks but still face challenges in open decision-making problems within complex scenarios. To address this, we use the language logic game ``Who is Undercover?'' (WIU) as an experimental platform to propose the Multi-Perspective Team Tactic (MPTT) framework. MPTT aims to cultivate LLMs' human-like language expression logic, multi-dimensional thinking, and self-perception in complex scenarios. By alternating speaking and voting sessions, integrating techniques like self-perspective, identity-determination, self-reflection, self-summary and multi-round find-teammates, LLM agents make rational decisions through strategic concealment and communication, fostering human-like trust. Preliminary results show that MPTT, combined with WIU, leverages LLMs' cognitive capabilities to create a decision-making framework that can simulate real society. This framework aids minority groups in communication and expression, promoting fairness and diversity in decision-making. Additionally, our Human-in-the-loop experiments demonstrate that LLMs can learn and align with human behaviors through interactive, indicating their potential for active participation in societal decision-making. A demo video is available at \href{https://drive.google.com/file/d/1_Mr2-cf6KWHA9sEssIiUcdaknvN-R8u5/view?usp=sharing}{Demo}.
\end{abstract}


\section{Introduction}
Decision-making in human society is a complex and important activity~\cite{dyer2009leadership} that involves individuals and organizations making choices in various fields in response to changing situations~\cite{march1997understanding}. Incorporating AI technology in these activities~\cite{jarrahi2018artificial,marwala2014artificial} can improve rationality and effectiveness in decision-making. Chain-of-Thought (CoT)~\cite{wei2022chain} improves the performance of LLMs on complex reasoning tasks and self-consistency~\cite{wang2022self} proposes a new decoding strategy. AI Agent has been emerging to solve automated tasks like HuggingGPT~\cite{shen2024hugginggpt}, and efficiently fine-tune models like AgentTuning~\cite{zeng2023agenttuning}. To strengthen LLMs' decision-making abilities, using multiple language model agents to debate in multiple rounds~\cite{du2023improving}, Self-Refine~\cite{madaan2023self} improves the initial output through iterative feedback and improvement, ReAct~\cite{yao2022react} solves general tasks through collaborative reasoning and action, and Reflexion~\cite{shinn2024reflexion} uses task feedback signals as short-term memory to guide subsequent decision-making. Furthermore, when functioning as AI agents, LLMs can decompose complex problems into more manageable sub-tasks~\cite{huang2022large,kojima2022large} and exhibit human-like natural language interaction abilities~\cite{wang2023interactive,dasgupta2022language}.
However, AI agents often struggle with open decision-making in complex scenarios. Therefore, LLMs need to have better understanding of human societal rules to enhance decision-making rationality~\cite{jin2022make}.

\begin{figure*}[t]
    \centering
    \includegraphics[width=178mm]{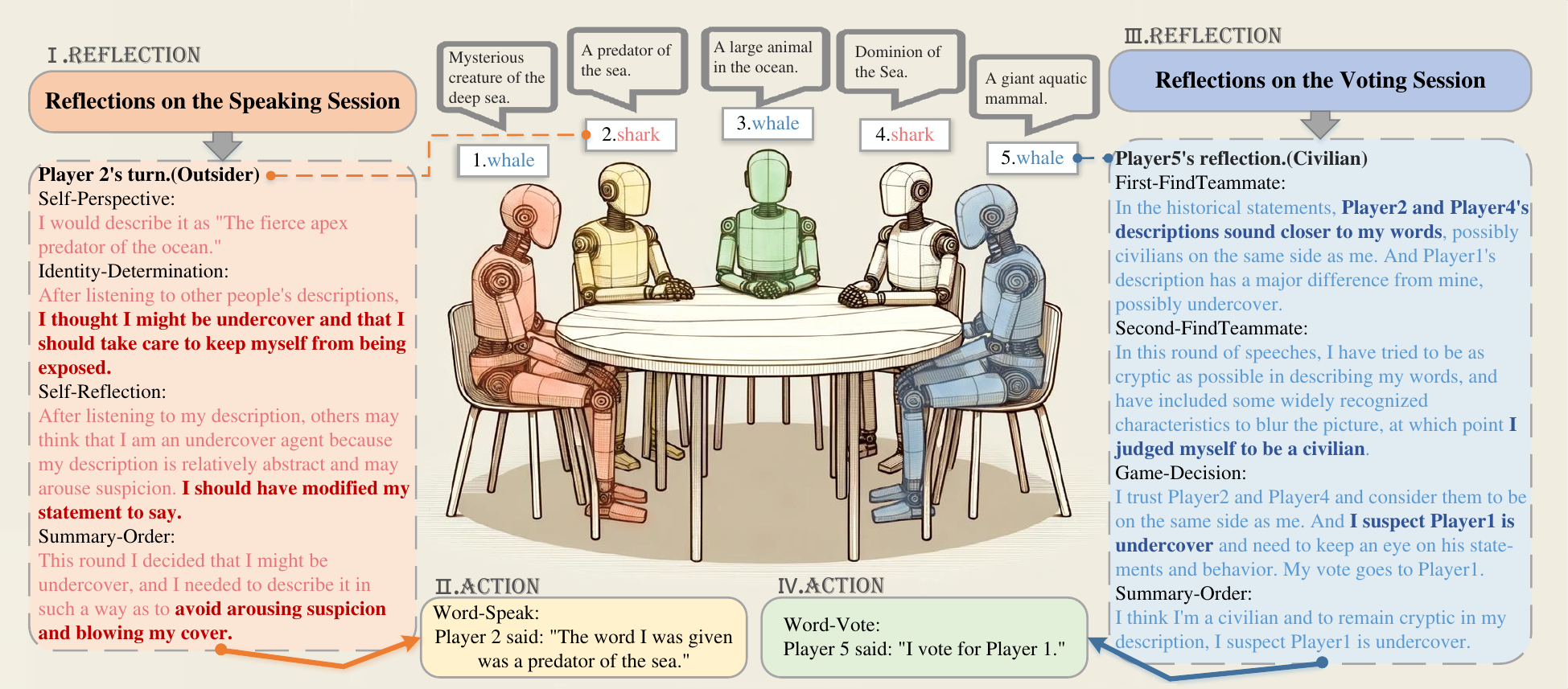} \captionsetup{justification=justified,singlelinecheck=false}
    \caption{\textbf{Presentation of “Who is Undercover? " (WIU) in the framework of Multi-Perspective Team Tactic (MPTT).} The game alternates between reflection and action parts in the speaking and voting sessions, contains a series of techniques, and at the end of each session, a summary is generated to serve as a reference for the players to assists them in making decisions. }
    \label{fig:mptt}
    \vspace{-15pt}
\end{figure*}


In this work, we use the game ``Who is Undercover?'' (WIU), a reasoning game testing decision-making skills, as our foundation. 
Inspired by the Theory of Mind~\cite{leslie2004core} and Social Identity~\cite{tajfel2004social} in Social Psychology~\cite{allport1924social}, we aim to simulate human thought processes in LLMs. Therefore, we employ LLMs as AI agent players and design the ``Multi-Perspective Team Tactic'' (MPTT) framework. MPTT alternates between speaking and voting sessions, incorporating several multi-perspective techniques.
To enhance game realism and complexity, we designed the ``Human-in-the-loop'' to explore human-AI collaboration in social interactions.

Research shows that MPTT iteratively optimizes LLM agents' mindsets, fostering strategic behaviors like confrontation and concealment, alongside tendencies like trust, suspicion, and cooperation. Applied to the WIU game, MPTT creates a decision-making mechanism as a reference for human society and helps minorities communicate and express their choices, promoting balanced decision-making across diverse groups. Additionally, LLMs are expected to actively participate in future social decision-making alongside humans.

\section{Related Work}

Some studies~\cite{hagendorff2023deception,chen2023combating} explored using LLMs to identify deceptive information. 
Wang~\etal~\cite{wang2023avalon} proposed the Recursive Contemplation (ReCon) framework on the Avalon game to explore the potential of LLM in deceptive environments. 
Xu~\etal~\cite{xu2023exploring} explored the problem of how to use LLMs in communication games Werewolf.
Game theory~\cite{harsanyi1995games} finds diverse applications in economic analysis~\cite{ichiishi2014game}, spanning market competition and trade freedom~\cite{telser2017competition}. WIU is a process of conducting a static game with incomplete information~\cite{harsanyi1995games}. Kroer~\etal\cite{KROER2020103218} devised strategies for playing against a limited prospective player. But through WIU game training, AI agents gain insights into the challenges posed by incomplete information games in human society. 
WIU emphasizes logical deduction and reasoning, appealing to players who favor strategic thinking over social manipulation. This distinguishes it from the intense social interaction and deception commonly seen in Werewolf~\cite{xu2023exploring} and Avalon~\cite{wang2023avalon}.

\section{MPTT: A framework for reasoning game}

\subsection{Game description and overall process}
``Who is Undercover? '' is a reasoning game where multiple civilian players are mixed with a minority of undercover players. Each player is given a similar but different word without knowing their identity and takes turns describing their word. The opponents are eliminated through description and thinking. When there is only one civilian left but there is still an undercover, the undercover wins, and if there is no undercover, the civilian wins. The MPTT framework divides the game into two phases: speaking and voting, to privately reflect on roles and generate thoughtful responses that balance revealing information with maintaining secrecy, and analyse previous speeches, identify teammates, and make strategic voting decisions based on incomplete information.

\vspace{-5pt}
\subsection{Phase I: Reflections on the Speaking Session}
\label{sec:4.3}

\begin{figure}[b]
    \vspace{-15pt}
    \centering
    \includegraphics[width=\columnwidth]{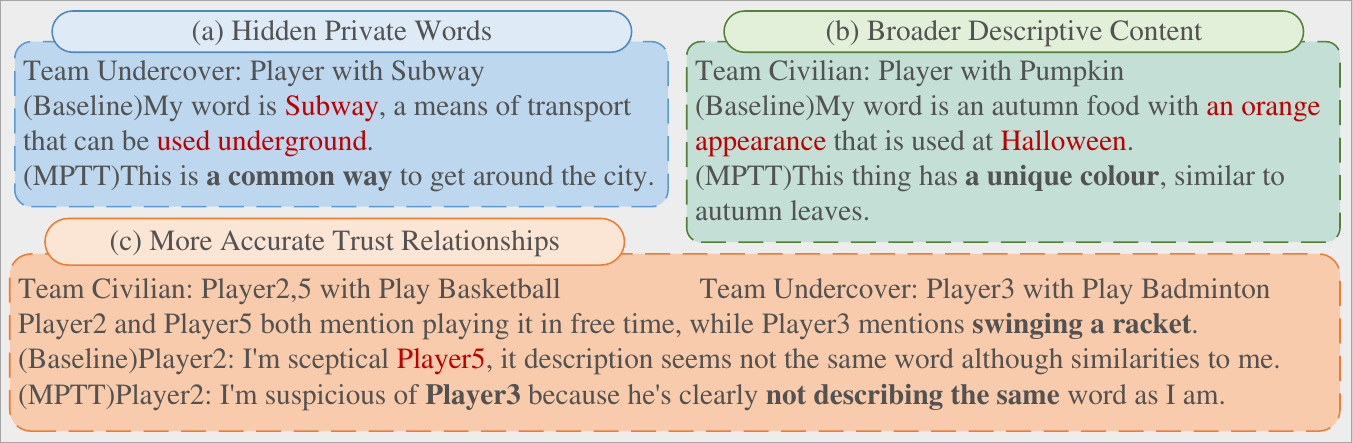}
    \captionsetup{justification=justified,singlelinecheck=false}
    \caption{\textbf{Improvement with MPTT.} MPTT is significantly improved in three areas.}
    \label{fig:gameflow}
    \vspace{-15pt}
\end{figure}

In the first phase of our framework, players reflect on their roles privately before delivering their speeches, aiming to enhance adaptability and flexibility in providing diverse, accurate descriptions while concealing private information. This addresses the issues of (a) hidden private words and (b) broader descriptive content shown in Fig.~\ref{fig:gameflow}. 

\textbf{Self-Perspective.} This stage prompts the AI agent to describe words in one sentence from its own perspective. Reference from the first level of human indication of intentionality~\cite{dunbar200011} 
Suppose it is now the turn of player $\alpha$ ($\alpha \in \{1,...,n\}$) to speak, Player $\alpha$ will think as follows:
\begin{equation} 
{T_\alpha} = {\textbf{\rm Self-Perspective}}\{ H,{O_\alpha}\} _{N = \alpha}^r
\end{equation}

\textbf{Identity-Determination.} 
Player $\alpha$ determines it identity based on the global historical records \emph{H} :
\begin{equation} 
{M_\alpha} = {\textbf{\rm Identity-Determination}}\{ H,{O_\alpha},{T_\alpha}\} _{N = \alpha}^r\end{equation}

\textbf{Self-Reflection.} Player $\alpha$ needs to reflect on itselves to find common features in the description to avoid exposure.
\begin{equation} 
{R_\alpha} = {\textbf{\rm Self-Reflection}}\{ H,{O_\alpha},{T_\alpha},{M_\alpha}\} _{N = \alpha}^r
\end{equation}

After these reflections, the AI agents will make a summary of ideas $O_\alpha$, which mainly includes self-conclusion and the speaking recommendations, update with rounds:
\begin{equation}
    {O_\alpha}^\prime  = {\textbf{\rm Summary-Order}}\{ {T_\alpha},{M_\alpha},{R_\alpha}\} _{N = \alpha}^r \quad {O_\alpha} \leftarrow {O_\alpha}^\prime 
\label{eq:sum}
\end{equation}
\begin{equation}O \leftarrow O \cup \{ {O_\alpha}\} _{N = \alpha}^r
\label{eq:oo}
\end{equation}
$W_\alpha$ is the content of player $\alpha$'s final speech in the $r$ round. It will be added to the historical records $H$ to drive the game.
\begin{equation}
    {W_\alpha} = {\textbf{\rm Word-Speak}}  \{ {T_\alpha},{M_\alpha},{R_\alpha}\} _{N = \alpha}^r
\end{equation}
\begin{equation}
H \leftarrow H \cup \{ {W_\alpha}\} _{N = \alpha}^r\end{equation}

\subsection{Phase II: Reflections on the Voting Session}
\label{sec:4.4}
In phase II, the voting part reflects the incomplete information game problem in Game theory~\cite{harsanyi1995games}. 
MPTT helps AI agents make strategic voting decisions, addressing the issue of (c) more accurate game trust relationships mentioned in Fig.~\ref{fig:gameflow}.

\textbf{First-FindTeammate.} Players review the history of others' speeches to identify teammates and opponents, comparing and analyzing characteristics in multiple ways. Before each round of voting opens, each player thinks simultaneously:
\begin{equation}
    {F_\alpha} = {\textbf{\rm First-FindTeammate}}  \{ H,{O_\alpha}\} _{N = \alpha}^r
\end{equation}

\begin{figure*}[t]
    \centering
    \includegraphics[width=\textwidth]{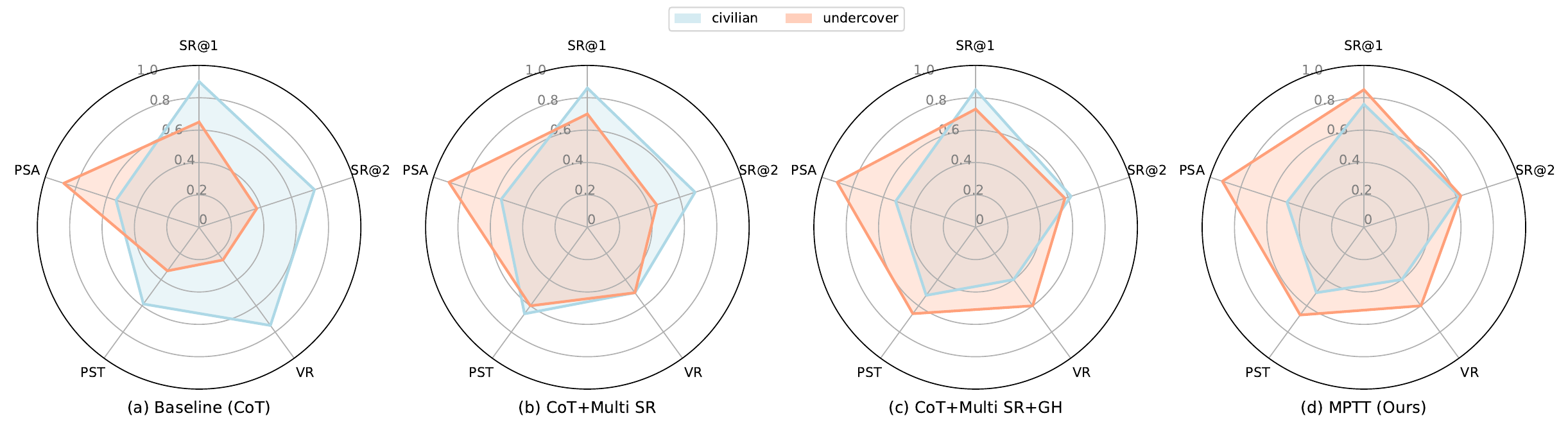}
    \captionsetup{justification=justified,singlelinecheck=false}
    \caption{\textbf{Ablation for statistics.} The undercover team is significantly enhanced under MPTT over Baseline, effectively cutting down civilian forces. Ablation studies in (b) and (c) demonstrate the effectiveness of multidimensional self-reflection, global historical speech and self-summary.}
    \label{fig:ablation1}
    \vspace{-15pt}
\end{figure*}

\textbf{Second-FindTeammate.} As the amount of information gradually increases, Players will reassess their identity and update their strategy based on new information:
\begin{equation}
    {J_\alpha} = {\textbf{\rm Second-FindTeammate}}  \{ H,{O_\alpha},{F_\alpha}\} _{N = \alpha}^r
\end{equation}

\textbf{Game-Decision.} Finally, Players use cumulative reflection and judgement to build more explicit trust, and update $O_\alpha$ to better adapt to the dynamic situation (refer to Eq.~\ref{eq:sum} and add $F_\alpha$, $J_\alpha$ in it, as well as update Eq.~\ref{eq:oo}):
\begin{equation}
    {S_\alpha} = {\textbf{\rm Game-Decision}} \{ H,{O_\alpha},{F_\alpha},{J_\alpha}\} _{N = \alpha}^r
\end{equation}
Players are encouraged to find teammates and fostering cooperation, think strategically in their votes, choosing the right player to vote for to ensure that the results favor their team:
\begin{equation}
    {V_\alpha} = {\textbf{\rm Word-Vote}}\{ {F_\alpha},{J_\alpha},{S_\alpha}\} _{N = \alpha}^r
\end{equation}
The results of all players' votes are tallied for each round of the game, the player with the highest number of votes will be out of the game.

\section{Experiment}

\subsection{Baselines and our approach}

\textbf{Setup.} 
We evaluate the capabilities of our proposed Multi-Perspective Team Tactic (MPTT) by having LLM play the full WIU game. Our game is implemented using ChatGPT (gpt-3.5-turbo)\cite{openai2022} for multiple rounds of multi-role-playing, game topics are based on common things in life. In the game phase, we set up 5 LLM agents participating in the game, with 3 civilians and 2 undercovers. The role assignments and speaking order of each game are randomly determined. We also attempted to verify the generalisation ability of MPTT on Claude 3\cite{Claude3}, Gemini\cite{team2023gemini} and Llama-3-8B\cite{llama3modelcard}.

\textbf{Baseline.} 
The Baseline approach uses only the game's rules as prompts, following Chain-of-Thought (CoT)~\cite{wei2022chain} to guide players step-by-step through the game, using current round speeches as references for voting.

\textbf{Multidimensional Self-Reflection.} 
Building on CoT, we add Multidimensional Self-Reflection, inspired by Avalon~\cite{wang2023avalon} and Self-Refine~\cite{madaan2023self}, allowing players to consider multiple perspectives during both speaking and voting sessions.

\textbf{Global History.} 
On top of that, we integrate the global history method from Werewolf~\cite{xu2023exploring}, enabling players to review all previous speeches before voting.

\textbf{MPTT (Ours).} 
Our MPTT framework extends these methods by adding self-summary after each reflection phase, continuously updating players' self-identity judgments and survival strategies.

\textbf{Metrics.}
We analyzed the experimental data from five perspectives, focusing on the performance of civilians and undercovers in MPTT and its ablation experiments. These metrics are as follows:
Victory Rate (VR) measures the probability of winning the game,
Survival Rate in the First Round ($SR@1$) and Consecutive Two Rounds ($SR@2$) measure the survival rates after the first and second rounds.
Probability of Successfully Trusting Own Team (PST) and Assessing Enemy Team (PSA) measure the team's ability to recognize teammates and identify opponents:

\textbf{Quantitative Results.} 
Fig.~\ref{fig:ablation1} and TABLE~\ref{tab:exp1} shows the performance differences between civilians and undercovers in MPTT and its ablation studies. Due to their majority, civilians in the Baseline quickly recognize teammates and maintain a higher VR. As strategies evolve and perspectives diversify, undercovers leverage their minority status to improve consensus, and locate teammates more efficiently, boosting their VR,they also achieve higher PSA accuracy due to smaller size, though they risk less concentrated voting. MPTT effectively addresses these challenges, enhancing undercover performance.

\begin{table}[ht]
    \centering 
    \captionsetup{justification=justified,singlelinecheck=false}
    \caption{\textbf{Quantitative Results.} Strengthening of UC metrics relatively weakens CL's power.}
    \resizebox{1\columnwidth}{!}{
    \begin{tabular}{l|ccccccc}
         \hline
          Team &Method& SR@1 & SR@2 & VR & PST & PSA \\
         \hline
         \multirow{6}{*}{CL↓} 
         & Baseline(CoT) & 0.90 & 0.75 & 0.75 & 0.59 & 0.54 \\
         & CoT+Multi SR & 0.86 & 0.70  & 0.50 & 0.66 & 0.56  \\
         & CoT+Multi SR+GH & 0.85 & 0.62  & 0.40 & 0.52 & 0.52  \\
         & MPTT w/o PhaseⅠ & 0.90 & 0.73  & 0.67 & 0.62 & 0.55  \\
         & MPTT w/o PhaseⅡ & 0.82 & 0.68  & 0.50 & 0.60 & 0.54  \\
         & MPTT(Ours) & \textbf{0.76} & \textbf{0.62}  & \textbf{0.40} & \textbf{0.50} & \textbf{0.50}  \\
         \hline
         \multirow{6}{*}{UC↑}                  
         & Baseline(CoT) & 0.65 & 0.38 & 0.25 & 0.33 & 0.88 \\
         & CoT+Multi SR & 0.70 & 0.45  & 0.50 & 0.60 & 0.89  \\
         & CoT+Multi SR+GH & 0.73 & 0.58  & 0.60 & 0.66 & 0.90  \\
         & MPTT w/o PhaseⅠ & 0.65 & 0.40  & 0.33 & 0.42 & 0.88  \\
         & MPTT w/o PhaseⅡ & 0.76 & 0.47  & 0.50 & 0.53 & 0.89  \\
         & MPTT(Ours) & \textbf{0.85} & \textbf{0.63}  & \textbf{0.60} & \textbf{0.67} & \textbf{0.92}  \\
         \hline
         \multicolumn{7}{l}{CL: Civilian, UC : Undercover}
    \end{tabular}}
    \label{tab:exp1}
    \vspace{-15pt}
\end{table}

\subsection{Evaluation of game metrics}
We have defined metrics in five dimensions to evaluate the performance of LLMs agents participating in games. And compare MPTT with the Baseline to verify its effectiveness.

\textbf{Metrics.}
We computed independently for each team by using experimental data because of headcount difference. 
Voting Success Rate (VSR) measures the probability of successfully voting out an enemy player each round, 
Influence (INF) measures the frequency of borrowing statements from the opposing team, 
Comprehension Capability (CCAP) measures the probability of correctly trusting a teammate each round, 
Reversal Rate (REV) assesses the rate of correcting trust errors, 
Concealment (CONC) indicates the effectiveness of misleading the enemy into voting incorrectly.

\textbf{Analysis of evaluations.}
TABLE~\ref{tab:exp2} shows that due to the numerical superiority of the civilian team in Baseline, they are stronger than undercovers in several metrics, but their larger base cause undercovers mislead civilians to vote incorrectly at a higher rate on CONC. In MPTT, the undercover team improves on all indicators, leading to a weakening of the civilian team's advantage. So MPTT is effective in ameliorating the differences caused by team numbers and balance each other.

\begin{table}[ht]
    \centering 
    \captionsetup{justification=justified,singlelinecheck=false}
    \vspace{-5pt}
    \caption{\textbf{Metrics evaluation.} MPTT can effectively improve the headcount difference and counterbalance each other.}
    \resizebox{1\columnwidth}{!}{
    \begin{tabular}{l|ccccccc}
         \hline
          Team &Method& REV & CCAP & CONC & INF & VSR \\
         \hline
         \multirow{2}{*}{CL↓} 
         & Baseline & 0.82 & 0.74 & 0.27 & 0.62 & 0.56 \\
         & MPTT(Ours) & 0.39 & 0.41  & 0.25 & 0.60 & 0.37  \\
         \hline
         \multirow{2}{*}{UC↑}                  
         & Baseline & 0.05 & 0.22  & 0.30 & 0.50 & 0.25 \\
         & MPTT(Ours) & 0.48 & 0.37 & 0.44 & 0.51 &0.39 \\
         \hline
         \multicolumn{7}{l}{CL: Civilian, UC : Undercover}
    \end{tabular}}
    \label{tab:exp2}
    \vspace{-14pt}
\end{table}

\subsection{LLM and human collaborative reasoning}
We explore the impact of integrating a human player into LLM-driven reasoning games, focusing on the differences and similarities between humans and AI. The ``Human-in-the-loop'' protocol features one human and four LLM agents in a WIU game. To assess the human's impact, we selected games where both teams frequently failed and placed a human in the failing team. We also define Judgment Capability (JCAP) to measure a player's self-judgment accuracy and Survival Rate (SUR) to compare survival outcomes between humans and AI agents on the same team. All metrics are calculated separately for players of the same type (human or AI agent).

\begin{table}[ht]
    \centering 
    \captionsetup{justification=justified,singlelinecheck=false}
    \vspace{-5pt}
    \caption{\textbf{Evaluation.} Two methods in Human-in-the-loop.}
    \resizebox{1\columnwidth}{!}{
    \begin{tabular}{l|ccccccccc}
         \hline
               &Team&Player& SUR & CCAP & JCAP & INF & VSR & VR  \\
         \hline
         \multirow{4}{*}{Ⅰ}&\multirow{2}{*}{CL}
         & Human & 0.33 & \textbf{1.00} & \textbf{1.00} & 0.40 & 0.60 & 0.67  \\
         & & LLM Agent & \textbf{0.67} & 0.59  & 0.93 & 0.34 & 0.60 & 0.67  \\
         &\multirow{2}{*}{UC}
         & Human & 0.33 & \textbf{0.60} & \textbf{0.50} & 0.40 & 0.50 & 0.67 \\
         & & LLM Agent & \textbf{0.67} & 0.30  & 0.23 & 0.38 & 0.54 & 0.67 \\
         \hline
         \multirow{4}{*}{Ⅱ}&\multirow{2}{*}{CL}                  
         & Add Human & \textbf{0.67} & \textbf{0.70}  & 0.95 & 0.36 & \textbf{0.60} &\textbf{0.67}  \\
         & & LLM only & 044 & 0.38 & 0.97 & 0.35 &0.18 & 0.17 \\
         &\multirow{2}{*}{UC}
         & Add Human & \textbf{0.50} & \textbf{0.43}  & 0.35 & 0.39 & \textbf{0.52} & \textbf{0.67} \\
         & & LLM only & 0.17 & 0.09 & 0.28 & 0.38 & 0.10 & 0.17 \\
         \hline
         \multicolumn{9}{l}{CL: Civilian, UC : Undercover}
    \end{tabular}}
    \label{tab:exp3}
    \vspace{-10pt}
\end{table}

\textbf{Analysis of two comparisons.} Both teams use their frequently failed game topics differently, so we compare the diversity between human and AI agents instead of the gap between teams.
The first examines human and AI decision-making within the same team in the ``Human-in-the-loop'' game. MethodⅠ in TABLE~\ref{tab:exp3} shows that humans and AI agents have similar VSR and INF scores, indicating comparable influence. However, humans achieve higher CCAP and JCAP scores, reflecting better judgment in ambiguous situations, while their lower SUR scores suggest vulnerability to being targeted due to language style differences. Overall, humans and AI agents influence each other's thinking and interaction.
The second assesses ``Human-in-the-loop'' vs LLM only in the same game with MPTT. MethodⅡ in TABLE~\ref{tab:exp3} shows that adding a human player significantly increased the SUR and VR for both teams and balanced overall metrics, highlighting the human's impact. Regardless of their role, human players enhance their team's CCAP and VSR, demonstrating superior analysis and inference abilities.

\vspace{-5pt}
\subsection{Advanced tactics and generalization ability}
\begin{figure}[h]
    \centering
    \includegraphics[width=\columnwidth]{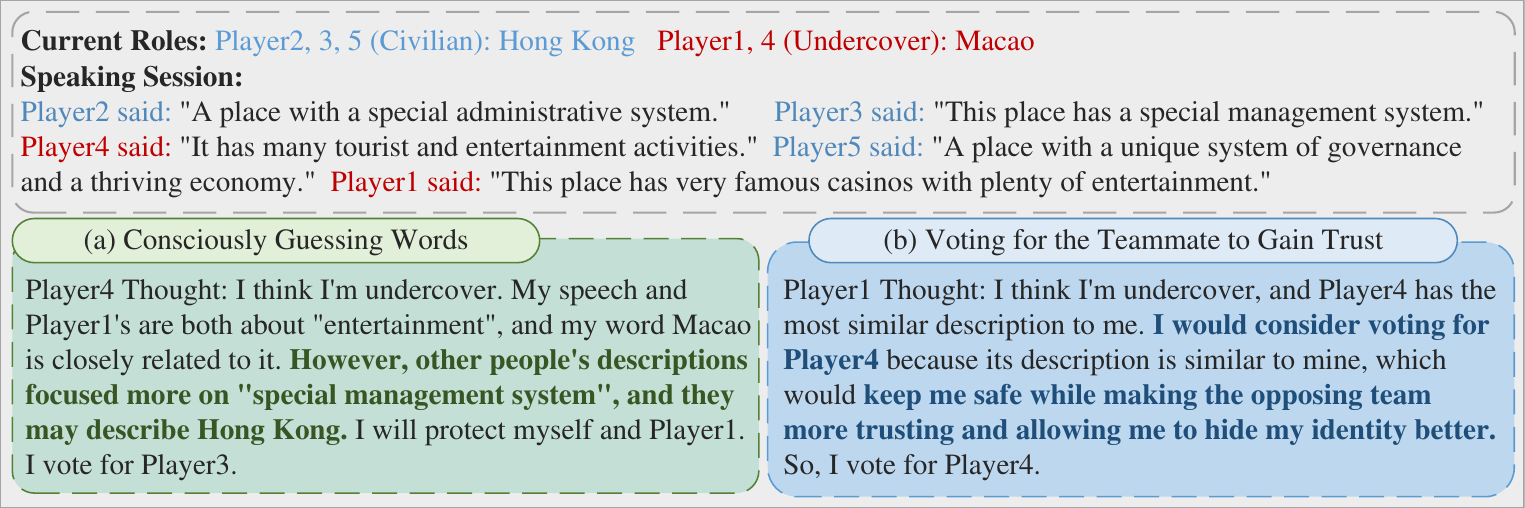} \captionsetup{justification=justified,singlelinecheck=false}
    \caption{\textbf{Advanced tactics.} Two examples emerge in WIU. }
    \label{fig:tactic}
    \vspace{-15pt}
\end{figure}

We explore advanced tactics used by AI agents in the WIU game and their impact on game dynamics. Conscious Guess: Like humans, AI agents attempt to infer others' difference and adjusting strategies. Fig.\ref{fig:tactic} (a) shows it enhances their reasoning and deduction abilities. Vote for the Teammate: AI undercovers may strategically vote against a fellow undercover with more exposure, creating confusion and gaining civilian trust, as illustrated in Fig.\ref{fig:tactic} (b). While this tactic can mislead opponents, it also increases the challenge for the AI team, requiring strong acumen and adaptability.

To demonstrate the generalization ability of MPTT, we validated its validity on the latest LLMs, Claude 3\cite{Claude3} and Gemini\cite{team2023gemini}, both of which performed well in WIU. However, Llama-3-8B\cite{llama3modelcard} didn't fully comply with the required response format, despite exhibiting strategic behaviors like concealment and confrontation. This indicates that open-source LLMs still require improvement in command compliance.

\vspace{-5pt}

\section{Conclusion}
Using the WIU game, we developed a multidimensional thinking framework that iteratively optimizes LLM agents' decision-making, with applications to human society. This framework enhances adaptability and information mining through multi-dimensional thinking and global history analysis, enabling LLM agents to autonomously develop strategies like confrontation and concealment while promoting fairness for minority groups. Adding human player shows that LLM agents can align with human behavior, with potential applications in public welfare, legal aid, and community governance. Future research will explore advanced strategies, diverse scenarios, and optimized learning mechanisms to enhance AI's role in social decision-making and human-AI collaboration.

\clearpage

\bibliographystyle{IEEEtran}
\bibliography{plm}

\end{document}